\documentclass[sigconf]{acmart}

\usepackage{luca}

\usepackage{subfigure}
\usepackage{comment} 
\newcommand{\diver}{\Delta}

\AtBeginDocument{%
  \providecommand\BibTeX{{%
    \normalfont B\kern-0.5em{\scshape i\kern-0.25em b}\kern-0.8em\TeX}}}

\setcopyright{rightsretained}
\copyrightyear{2021}
\acmYear{2021}
\acmDOI{}

\acmConference[Responsible AI @ KDD 2021]{KDD Workshop on Measures and Best Practices for Responsible AI - Responsible AI @ KDD 2021}{August 14-18, 2021}{Virtual Event}
\acmBooktitle{Workshop on Measures and Best Practices for  Responsible AI at ACM KDD 2021 (Responsible AI @ KDD 2021)}
\acmPrice{}
\acmISBN{}

\begin{document}

\sloppy

\title{Identifying Biased Subgroups in Ranking and Classification}


\author{Eliana Pastor}
\orcid{0000-0002-3664-4137}
\email{eliana.pastor@polito.it}
\affiliation{%
  \institution{Politecnico di Torino, Italy}
  \country{}
}

\author{Luca de Alfaro}
\email{luca@ucsc.edu}
\orcid{0000-0003-3856-4576}
\affiliation{%
  \institution{UC Santa Cruz, USA}
  \country{}
}

\author{Elena Baralis}
\orcid{0000-0001-9231-467X}
\email{elena.baralis@polito.it}
\affiliation{%
  \institution{Politecnico di Torino, Italy}
  \country{}
}


\begin{abstract}

When analyzing the behavior of machine learning algorithms, it is important to identify specific data subgroups for which the considered algorithm shows different performance with respect to the entire dataset. The intervention of domain experts is normally required to identify relevant attributes that define these subgroups. 

We introduce the notion of divergence to measure this performance difference and we exploit it in the context of (i)~classification models and (ii)~ranking applications to automatically detect data subgroups showing a significant deviation in their behavior. Furthermore, we quantify the contribution of all attributes in the data subgroup to the divergent behavior by means of Shapley values, thus allowing the identification of the most impacting attributes.

\end{abstract}



\newcommand{\compas}{{\sc compas}}
\newcommand{\attr}{\textrm{attr}}
\newcommand{\itemsets}{\mathcal{I}}
\newcommand{\statistic}{\mathbf{G}}
\newcommand{\su}{\Delta^g} 
\newcommand{\asu}{\widetilde{\Delta}^g}
\newcommand{\support}{\mathcal{S}}
\newcommand{\globunf}{TODO?}
\newcommand{\falsepos}{\textit{fp}}
\newcommand{\falseneg}{\textit{fn}}
\newcommand{\truev}{o_t}
\newcommand{\predv}{o_p}
\newcommand{\dataset}{D}
\newcommand{\domain}{\mathcal{D}}
\newcommand{\Beta}{\mathrm{Beta}}
\newcommand{\prob}{\mathrm{Pr}}
\newcommand{\minsup}{s}


\maketitle

\begin{table}
\begin{center}
\setlength\tabcolsep{2.15pt} 
\begin{tabular}{lrrrr}
\toprule
Data subgroup &   FPR &   FNR &  $\Delta_{FPR}$ &  $\Delta_{FNR}$  \\
\midrule
  Entire dataset  &  0.09 &  0.70 &  0.00 &  0.00 \\
  age<25, \#prior>3, sex=Male &  0.68 &  0.36 &  0.59 & -0.34 \\
 age>45, charge=M, race=Cauc &  0.01 &  1.00 & -0.08 &  0.30 \\
 \bottomrule
\end{tabular}
\end{center}
\caption{Example of patterns in the \compas\ dataset, along with their false-positive rate (FPR), false-negative rates (FNR) and divergence $\Delta$. 
}
\label{table-teaser}
\end{table}

\section{Introduction}

Machine learning models and automated-decision making procedures are becoming more and more pervasive, thus a growing interest is arising on the careful understanding of their behavior~\cite{surveyxaifosca, zehlike2021fairness}. A relevant step in the explanation of the outcome of machine learning algorithms is the identification of data subgroups in which the considered algorithm may show a different, and potentially anomalous, behavior. The identification of peculiar behaviors of data subgroups finds important applications in the KDD pipeline, ranging from model validation and testing~\cite{ chung2019automated, wu2019errudite} to the evaluation of model fairness~\cite{chung2019automated, cabrera2019fairvis}. In particular, societal bias~\cite{barocas2016big} is becoming a growing concern and researchers are increasingly working on measuring and ensuring fairness in machine learning. To this aim, human experts are frequently required to manually identify sensitive attributes (e.g., gender, ethnicity) or problematic data subgroups.

The behavior of machine learning algorithms is frequently evaluated by means of global metrics, which consider the performance of the algorithm on a global level, for the entire dataset, or for specific class labels. Differently, in this paper we propose the concept of \textit{divergence} as a measure of the difference in statistics (e.g., false positive rate) between the behavior of the machine learning algorithm on a data subgroup and on the entire dataset. Data subgroups showing a significant deviation in their behavior are automatically identified by our approach, which characterizes data subgroups by means of a combination of attribute values, denoted in the paper as \textit{patterns}, or \textit{itemsets}.

To illustrate the concept of divergence, consider the \compas\ dataset~\cite{angwin_2016}, in which a score measuring recidivism risk is assigned to criminal defendants by a proprietary algorithm. In this case, the positive class corresponds to high recidivism scores.
Table~\ref{table-teaser} shows the false-positive (FPR) and false-negative (FNR) rates occurring in the entire dataset (first row) and in the data subgroups characterized by the highest FPR (second row) and FNR (third row) divergence. For example, the pattern (\textit{age>45, charge=M, race=Cauc}) shows a high $\Delta_{FNR}$ divergence. Hence, instances belonging to this data subgroup will be wrongly assigned to the negative class with a higher rate with respect to the entire dataset.

We propose a general framework for divergence computation, that allows the automatic identification of problematic data subgroups both in classification and ranking problems. Our approach builds on well-known itemset mining algorithms (e.g., FP-growth~\cite{han2000mining}) and allows the efficient identification of all the problematic patterns above a (low) frequency threshold. Moreover, given a specific pattern, we exploit the notion of Shapley value~\cite{shapley1953value} to quantify the contribution of each attribute value to the pattern divergence. For example, as will be shown in the following sections, the attribute value \textit {age>45} is contributing the most to the divergent behavior of the pattern discussed before.

\section{Related Work}

Existing techniques for analyzing data subgroups include both supervised and unsupervised techniques. 
Supervised techniques rely on domain experts and users to identify the subgroups of interest.
Several tools analyze performance over data subgroups specified by the user of a classification model for validation purposes~\cite{tfma_medium, kahng2016visual}.
Many efforts have been devoted to detecting and mitigating bias in classification tasks.
Several approaches evaluate if different treatment or performance occur on groups determined by some sensitive or protected attributes~\cite{foulds2020intersectional, dwork2018group, kearns2018preventing, morina2019auditing}.
Recently, researches focused the attention on fairness in rankings~\cite{zehlike2021fairness}.  
Different works propose measures and mechanisms to audit ranking outputs and mitigate bias over protected groups~\cite{yang2017measuring,zehlike2017fa, zehlike2020reducing, celis2017ranking}.
The analysis of group fairness in both the classification and the ranking tasks generally assumes that the sensitive attributes (e.g. sex, race, age, degree of disability) that define the protected groups  are known or specified a priori. 
We propose an approach for the automatic identification of critical subgroups treated differently by a generic model, be it a classifier or a ranker,  without the a priori knowledge of the groups and attributes of interest. We concern ourselves on auditing differences in subgroups, rather than mitigation strategies, which may be application-dependent.

Several works have been proposed to automatically identify critical subgroups in the classification domain~\cite{cabrera2019fairvis, wu2019errudite, chung2019automated}. 
FairVIS~\cite{cabrera2019fairvis} audits the fairness of classification models leveraging on a clustering-based subgroup identification technique. 
Fairness and performance metrics are evaluated on the identified clusters described by a few dominant features obtained via feature entropy. 
Differently, we exploit frequent pattern mining algorithms to identify frequent critical subgroups. 
The subgroups are obtained by slicing the attribute domains. Hence, the characterizing features are known and readily interpretable.
Errudite~\cite{wu2019errudite} is an interactive system that enables data grouping for NLP error analysis using a domain-specific language.
Differently from~\cite{wu2019errudite}, our approach deals with structured data and slices the data by (discrete) attribute values. 

Slice Finder~\cite{chung2019automated, slicefindericde} automatically detects data slices in which a classification model performs poorly. Similarly to our approach, the data subgroups are identified by slicing via attribute values.
Slice Finder defines a top-down lattice search to find the top-k critical slices. The data exploration, based on breadth-first traversal, stops when it reaches a subgroup characterized by a sufficiently large difference in performance which is statistically significant. 
However, this stopping criterion may prevent finding relevant critical subgroups, because the metrics used for assessing model performance on subgroups are typically non-monotone. 
Thus, from the critical behavior of a group, we cannot make assumptions on the behavior of its super/sub-groups. 
We propose a more comprehensive exploration by identifying all the subgroups adequately represented (i.e., above a frequency threshold) in the dataset. We then characterize the subgroups using the notion of Shapley values~\cite{shapley1953value} to estimate the contribution of each attribute value to the subgroup divergence.

We introduced the notion of divergence, and the use of Shapley value to measure the contribution of attributes to divergence, in \cite{divexplorerpaper}.  
This paper extends the definitions to handle ranking systems, as well as general quantitative prediction functions. 


\section{Example Datasets}

As running examples to illustrate the concepts, we use two well-known datasets.
The first is the \compas\ dataset~\cite{angwin_2016}, which consists of defendants considered for release on parole.
For each individual, the dataset contains personal data such as age range, race, gender, and data related to criminal history, such as number of prior offenses. 
The \compas\ dataset contains also a score that estimates the likelihood that a defendant commits another offense (recidivates) in the next two years. From this score, via comparison with a threshold, we can obtain a binary classification. 
For each defendant, the ground truth is known, so that the false-positive and false-negative rates of the classification can be computed. 
We are interested in characterizing the subgroups for which these rates deviate from the average. 

The second dataset we consider is the Law School Dataset. 
The Law School Admission Council conducted a survey across 163 law schools in the United States in 1998 \cite{wightman98}. 
The resulting Law School Dataset contains information on 21,791 law students such as their entrance exam scores (LSAT), their grade-point average (GPA) collected prior to law school, and their normalized first year average grade (ZFYA), in addition to their race and sex. 
We use the dataset as prepared by \cite{kusner17}.
In this dataset, we study how the average ZFYA score, and the average rank of students after the first year, vary across subgroups. 

\section{Divergence}

We provide here the definition of \emph{divergence}, which captures the difference between statistical measures computed on individual subgroups, versus the entire dataset, and we illustrate the divergence of various measures on our example datasets.  

\subsection{Datasets and itemsets}

We consider datasets $\dataset$ in tabular form: there is a fixed set $A$ of columns, and a set $X$ of rows;  every row $x$ assigns value $x(a)$ to attribute $a \in A$. 
Thus, $A$ is the schema of the dataset, and the rows $X$ are the \emph{instances}.
We assume that every attribute $a \in A$ has a finite domain $D_a$. Thus, the dataset is discretized. For example, in our \compas\ dataset, an instance is a defendant, and the attributes are age range, gender, and so on. 

An \emph{item} $\alpha$ consists in a selection $a=c$ for an attribute $a\in A$ and a value $c \in D_a$. 
An \emph{itemset} is a set of items, with each item involving a distinct attribute. 
For instance, in \compas, an itemset is \{\textit{age>45, race=Caucasian}\}.
The \emph{support-set} $\dataset(I) = \set{x \in \dataset \mid x \sat I}$ of an itemset $I$ consists of the instances that satisfy $I$; the \emph{support} of $I$ is the fraction of dataset instances in $\dataset(I)$, or $\sup(I) = \frac{| \dataset(I) |}{| \dataset |}$. 

\subsection{Itemset divergence} 

We are interested in identifying subgroups of data that behave differently, compared to the overall dataset, with respect to statistical measures.
For instance, in classifiers we are interested in identifying data subgroups where the false-positive and false-negative rates differ from the average. In rankings, we are interested in data subgroups where the average rank deviates from the global one. 
We use itemsets to describe data subgroups, and we use an \emph{outcome function} to capture the statistic of interest. 

An \emph{outcome function} $o: X \mapsto \set{\bot} \union \reals$  associates with each instance either a do-not-consider value $\bot$, or a real number. 
For a (possibly empty) dataset $I$, we define the \emph{outcome $o(I)$ on $I$} via: 
\begin{equation} \label{eq-outcome}
    o(I) = E \set{o(x) \mid x \sat I, o(x) \neq \bot} \; . 
\end{equation}
If $I$ is empty, $o(\emptyset)$ is the outcome of the complete dataset. 
We then define the \emph{divergence} of $I$ (with respect to outcome function $o$) to be: 
\begin{equation} \label{eq-divergence}
    \Delta_o(I) = o(I) - o(\emptyset) \; . 
\end{equation}
The divergence of an itemset captures the difference in behavior between the itemset, and the entire dataset, with respect to the outcome function under consideration. 
We illustrate, via examples, how different outcome functions enable the analysis of raw datasets, as well as the behavior of classifiers and ranking systems. 

\subsection{Attribute divergence}

\begin{table}
\setlength\tabcolsep{1.15pt} 
\begin{tabular}{lrrc}
\toprule
Itemset &   Sup &  $\Delta_{ZFYA}$  & $t$\\
\midrule
LSAT>41.0, UGPA>3.5, race=White, sex=Female &  0.03 &     0.4115 &       11.1 \\
LSAT>41.0, UGPA>3.5, race=White &  0.07 &     0.4063 & 16.8\\
LSAT>41.0, UGPA>3.5, race=White, sex=Male &  0.04 &     0.4025 & 13.0\\
\midrule
LSAT$\leq$33.0, race=Black, sex=Male &  0.02 &    -1.0257 & 21.2 \\
LSAT$\leq$33.0, UGPA$\leq$3.0, race=Black, sex=Male &  0.01 &    -1.0049 & 17.05\\
LSAT$\leq$33.0, race=Black &  0.05 &    -0.9787 & 33.3\\
\bottomrule
\end{tabular}
\caption{Top-3 itemsets with highest and lowest ZFYA divergence for the Law School Dataset. The support threshold is $s=0.005$.}
\label{table-law-score-zfya}
\end{table}

In many cases, one can take the outcome of an instance to be one of the quantitative attributes of the instance itself. 
For the Law School Dataset, the simplest choice consists in taking $o(x) = ZFYA(x)$, setting the outcome equal to the normalized first-year average of each student. 
Table~\ref{table-law-score-zfya} lists the three itemsets with greatest positive and negative divergence, among those with support at least $s=0.005$, which corresponds to about 100 students. 
The table reports also the $t$-value of the divergence, computed according to Welch's $t$-test. 
We use a support limit both to provide a termination criterion for the divergence computation algorithm, as discussed in Section~\ref{sec-algo}, and to exclude itemsets with such small support that the analysis is affected by statistical fluctuations. 
From the results, we see that the itemset with greatest positive divergence is \{LSAT>41.0, UGPA>3.5, race=White, sex=Female\}, for which the ZFYA-divergence is 0.41. 
The itemset with the greatest negative divergence is \{LSAT$\leq$33.0, race=Black, sex=Male\}, for which the ZFYA score is on average lower by 1.03 compared with the dataset average. 
In Section~\ref{sec-contribution}, we will see how to analyze the contribution of each of the three items LSAT$\leq$33.0, race=Black, and sex=Male, to the divergence of this itemset. 

\subsection{Classifier divergence}

Divergence can also be applied to analyze classifier behavior.
Given a classifier, let $p(x) \in \set{\true, \false}$ be the predicted value for an instance $x$, and let $t(x)$ be the true value (ground truth). 
In a classifier, it is often of interest to study the divergence of the false-positive rate (FPR) and false-negative rate (FNR). 
The variation of these rates across data subgroups gives an indication of how the subgroups are advantaged, or disadvantaged, by classifier errors. 
To capture the divergence of the false-positive rate, we use the outcome function: 
\[
  o(x) = \begin{cases}
    \bot & \mbox{if $t(x) = \true$} \\
    0 & \mbox{if $t(x) = \false$ and $p(x) = \false$} \\
    1 & \mbox{if $t(x) = \false$ and $p(x) = \true$} 
  \end{cases}
\]
for $x \in X$. 
Here, the outcome $\bot$ is used to exclude from the statistic the true-positives, so that the outcome $o(I)$ of an itemset $I$ is its FPR.
Outcome functions for capturing the FNR, true-positive rate, and so on, can be similarly defined. 

\begin{table}
\setlength\tabcolsep{2pt} 
\begin{tabular}{lllc}
\toprule
 Itemset &   Sup &       $\Delta_{FPR}$ &     $t$ \\
\midrule
 age<25, \#prior>3, sex=Male &  0.02 &  0.594 &  6.1 \\
 age<25, \#prior>3 &  0.02 &  0.527 &  5.7 \\
 age<25, stay=1w-3M, race=Afr-Am, sex=Male &  0.02 &  0.306 &  3.8 \\
\bottomrule 
   & Sup &  $\Delta_{FNR}$ &  $t$ \\
 \hline
 age>45, charge=M, race=Cauc &  0.05 &  0.302 &  17.6 \\
 age>45, charge=M, \#prior=0 &  0.04 &  0.302 &  10.4 \\
 age>45, charge=M, \#prior=[1,3] &  0.03 &  0.302 &  14.1 \\
\bottomrule 
\end{tabular}
\caption{Top-3 divergent patterns with respect to FPR and FNR for the \compas\ dataset. The support threshold is $s=0.0175$.}
\label{table-compas-divergence}
\end{table}

In Table~\ref{table-compas-divergence} we report the top-3 divergent patterns with respect to FPR and FNR, for a minimum support of $s=0.0175$, equivalent to about 100 instances.
An itemset with positive divergence for FPR is an itemset consisting of defendants that are incorrectly predicted to recidivate at a rate higher than the average for the dataset.
We see that young males, with prior crimes, are the defendants most often falsely predicted to recidivate. 
Conversely, old Caucasian males are the most frequent instances incorrectly predicted \emph{not} to recidivate. 

\subsection{Ranking divergence}

In a ranking system, every instance $x$ has a \emph{rank} $i(x) \in \naturals_{>0}$, where $i=1$ is the top rank. 
It is natural to define the outcome function $o$ via a \emph{rank valuation function} $\gamma: \naturals_{>0} \mapsto \reals$, where $\gamma(i)$ represents the value, to an instance, of being ranked in position $i$. 
We define the outcome of instance $x \in X$ via: 
\begin{equation}
    o(x) = \gamma(r(x))
    \label{eq-gamma-i}
\end{equation}
The outcome $o(I)$ of an itemset $I$ would then correspond to the average value an instance in $I$ receives from being ranked. 

As an example, consider admissions to a university. 
If applicants are ranked, and the top $k$ admitted, we can take $\gamma(i) = 1$ for $i \leq k$ and $\gamma(i) = 0$ otherwise. 
The outcome $o(I)$ corresponds to the admission rate of $I$, that is, the fraction of applicants in $I$ that are admitted, and the divergence $\Delta_o(I)$ would then represent how more, or less, likely applicants in $I$ are to be admitted, compared with the general population. 
Notice that the use of a rank-value function $\gamma$, rather than simply the rank, is key to capturing the impact of the ranking on instances in top-$k$ admissions.

\begin{table}
\setlength\tabcolsep{1.15pt} 
\begin{tabular}{lrrc}
\toprule
Itemset &   Sup &  $\Delta_{\gamma}$ & $t$\\
\midrule
 LSAT>41.0, UGPA>3.5, race=White, sex=Female &  0.03 &     0.0206 &        8.7 \\
             LSAT>41.0, UGPA>3.5, race=White &  0.07 &     0.0196 &       13.0 \\
   LSAT>41.0, UGPA>3.5, race=White, sex=Male &  0.04 &     0.0189 &        9.9 \\
                         \midrule
LSAT$\leq$	33.0, race=Black, sex=Male &  0.02 &    -0.0283 &       25.6 \\
   LSAT$\leq$	33.0, UGPA$\leq$	3.0, race=Black, sex=Male &  0.01 &    -0.0280  &       21.0 \\
             LSAT$\leq$	33.0, UGPA$\leq$	3.0, race=Black &  0.03 &    -0.0278 &       31.4 \\
\bottomrule
\end{tabular}
\caption{Top-3 itemsets with highest and lowest divergence for the Law School Dataset, for the internship example, with $\gamma(i) = i^{-0.1}$. The support threshold is $s=0.005$.}
\label{table-law-beta}
\end{table}

As another example, returning to our Law School Dataset, assume that at the end of their first year, students internship applications are displayed to internship hosts sorted according to the first-year average grade (ZFYA) of a student.  
Assume that the benefit of being ranked in position $i$ to the student is proportional to $\gamma(i) = i^{-0.1}$; this type of relation between rank and benefit is common in search applications. 
Table~\ref{table-law-beta} gives the itemsets with top and bottom divergence with respect to this benefit. 
We see that the itemset that derives the most benefit out of internships would be \{LSAT>41.0, UGPA>3.5, race=White, sex=Female\}, and the one deriving the least benefit would be \{LSAT$\leq$33.0, race=Black, sex=Male\}.

\subsection{Computing the divergence}
\label{sec-algo}

We compute the outcome (\ref{eq-outcome}) of itemsets, and hence their divergence, by extending frequent-pattern mining algorithms \cite{10.5555/3208440,agrawal1994fast,han2000mining}.

The input to the algorithms is a \emph{support threshold} $s$: only itemsets $I$ with $\sup(I) \geq s$ are considered.
The support threshold is chosen according to the minimum size of the data subgroup that one is interested in investigating.
For each itemset $I$ above the support threshold, its outcome $o(I)$ is computed by tallying, during itemset extraction, (a) the sum of all instance outcomes in $I$ that are not $\bot$, and (b) the number of such instances, and then taking the ratio. 

By extending frequent-pattern mining algorithms, we obtain algorithms that compute the divergence of all itemsets above the support size extremely efficiently. 
Even when there are $10^5$ itemsets above the support threshold, they can be identified and their divergence computed in a matter of a dozen seconds or so. 

\section{Item Contribution to Divergence}
\label{sec-contribution}

Once one has identified the itemsets for which the relevant classification or rank statistics most deviate from the average, it is of interest to determine the contributions of the individual items to the divergence of the itemset. 
For instance, we see from Table~\ref{table-compas-divergence} that the defendants most likely to be incorrectly predicted to recidivate are those in \{age<25, \#prior>3, sex=Male\}; which one of the three items age<25, \#prior>3, and sex=Male, is the most important?
Of which fraction of the divergence $0.594$ is each item responsible?

To answer this question, we use the notion of Shapley values \cite{shapley1953value}. 
In game theory, Shapley values are a way of distributing the value $v(1, 2, \ldots, N)$ of a team of $N$ players, to each of the players $1, 2, \ldots, N$, in such a way that the sum of the player's values is equal to the team's value. 
We define the contribution $\Delta(\alpha \mid I)$ of item $I$ to the divergence $\Delta(I)$ as the Shapley value of $\alpha$ in $I$, for value $\Delta$:  
\begin{equation} \label{eq-individual}
  \Delta(\alpha \mid I) = 
  \sum_{J \subs I \setminus \set{\alpha}} 
    \frac{|J|! (|I| - |J| - 1)!}{|I|!}
    \bigl[ \diver(J \union \alpha) - \diver(J) \bigr] \: .
\end{equation}	
Note that, if an itemset is above the support threshold, all its subsets also are, so all divergences of itemsets in (\ref{eq-individual}) can be computed by our algorithm.

\begin{figure}
    \includegraphics[width=0.99\linewidth]{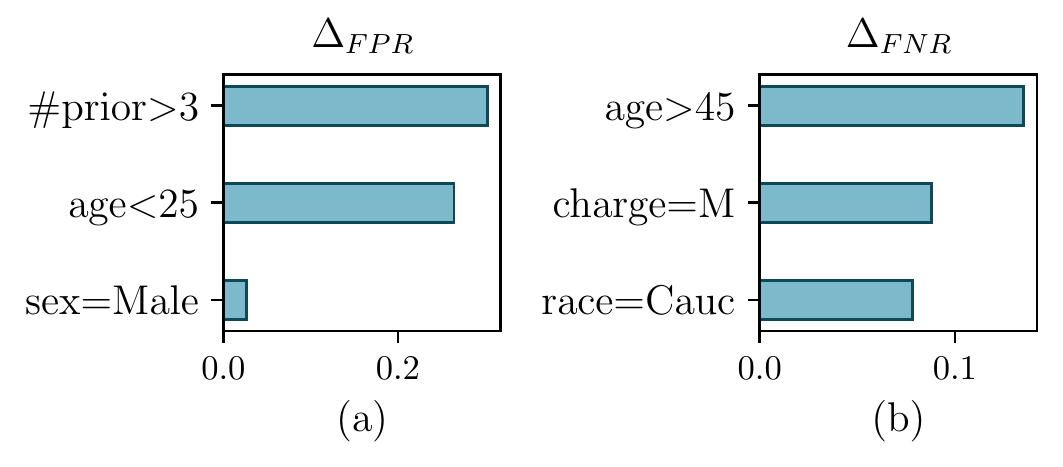}
    \caption{Contributions of individual items to the divergence of the \compas\ frequent patterns having greatest FPR and FNR divergence in Table~\ref{table-compas-divergence}.}.
    \label{fig-compas-shapley-local}
\end{figure}

\begin{figure}
    \includegraphics[width=0.55\linewidth]{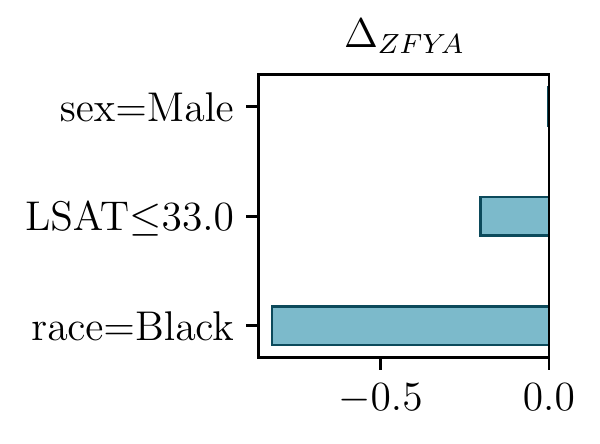}
    \caption{Contributions of individual items to the divergence of the frequent patterns having lowest ZFYA divergence for the  Law School Dataset ($s=0.005$).}
    \label{fig-law-zfya-shapley-local}
\end{figure}

Figure~\ref{fig-compas-shapley-local} reports the influence of individual items to the itemsets with greatest FPR and FNR divergence.
We see that sex=Male is responsible only for a small fraction of the FPR divergence of \{age<25, \#prior>3, sex=Male\}, while age<25 and \#prior>3 have effects of similar magnitude. 

Figure~\ref{fig-law-zfya-shapley-local} reports the results of a similar analysis for the itemset with lowest ZFYA divergence in the Law School Dataset.
We see that race=Black is the predominant factor, with LSAT$\leq$33.0 giving a minor contribution, and sex=Male a negligible one. 
The predominant role of race stands out as a warning signal, indicating that this negative rank divergence merits further investigation. 

\section{Conclusions}

In this paper we presented a method for finding data subgroups that behave differently from the overall dataset in classification, ranking, or other automated prediction systems. 
At the core of the approach is the notion of \emph{divergence}, which quantifies the difference in behavior, and which can be efficiently determined via data mining approaches. 
We believe our approach may provide a useful building block in strategies to mitigate algorithmic bias. 

\begin{acks}
This work has been partially supported by the SmartData@PoliTO center on Big Data and Data Science.
\end{acks}

\bibliographystyle{ACM-Reference-Format}
\balance
\bibliography{divexplorer.bib}

\end{document}